\documentclass[letterpaper]{article}
\usepackage{aaai}
\usepackage{times}
\usepackage{helvet}
\usepackage{courier}
\usepackage{graphicx}
\usepackage{url}
\begin{document}
%
\title{Rapid Prediction of Player Retention in 
Free-to-Play Mobile Games}
\author{
Anders Drachen\\
Eric Thurston Lundquist\\
Yungjen Kung\\
Pranav Simha Rao\\
Diego Klabjan\\
Rafet Sifa\\
Julian Runge
}
\maketitle
\begin{abstract}
\begin{quote}
Predicting and improving player retention is crucial to the success of mobile Free-to-Play games. This paper explores the problem of rapid retention prediction in this context. Heuristic modeling approaches are introduced as a way of building simple rules for predicting short-term retention. Compared to common classification algorithms, our heuristic-based approach achieves reasonable and comparable performance using information from the first session, day, and week of player activity.
\end{quote}
\end{abstract}

\section{Introduction}
Predictive modeling in Free-to-Play (F2P) games has become a regular occurrence in the mobile gaming industry as well as within the associated academic fields investigating player behavior at large scales. 
Previous work has seen the development of a variety of machine learning-based models \cite{Runge2014-CPH,Sifa2015-PPD,Hadiji2014-PPC,Nasr2013-GAM,pittman2010-CVP,thawonmas2011-ARO,Mahlmann2010-PPB,yang2014-ipc,Xie2015-PPD}, and has focused on situations where there is at least a week or even more data available about the players \cite{Hadiji2014-PPC,Sifa2015-PPD,runge2014-PAS}. However, in a commercial context there is a direct interest in being able to predict player retention as fast as possible. There are multiple reasons for this, but one of the primary ones is that F2P games generally lose a majority of the players to churn within the first few days after an install \cite{nozhnin2013-PCW,Runge2014-CPH,rothenbuehler2015-HMM}. Predictions are also important for appropriate incentivization of players to remain in the game \cite{Runge2014-CPH}. Essentially, there are two steps in solving the problem of players leaving a game: 1) Predicting if a player will churn or not, and when; 2) Identifying how to prevent this from happening or, if not deemed possible, recommending a different suitable game to the player. The earlier a correct prediction can be made after a player starts playing a new game, the more valuable that knowledge will be. Fast predictions enable companies to build well tailored customer relationship management and respond to user behavior proactively \cite{runge2014-PAS,Sifa2015-PPD,rothenbuehler2015-HMM,Xie2015-PPD}. 

As many companies in the mobile gaming industry have rather small operations, they cannot afford their own in-house analytics. It is thus imperative to identify simple, frugal, but effective prediction models to make the benefits of predictive analytics accessible to them. But heuristic models also bear value regardless of company size and cash balance. Especially when a game is freshly launched and there is an overly full pipeline of features to be built, reducing a complex predictive effort to an easily implementable decision rule is of value to large and small companies alike. To address this challenge we introduce the idea of \emph{heuristic} modeling and forecasting \cite{Goldstein2011-FAF,Gigerenzer2009-HHW,Artinger2015-HAD}. Heuristics are simple, computationally fast and robust rule systems that are often derived from intuition or a combination of intuition and data-driven modelling. They are potentially beneficial along several dimensions: a) They are easy to deploy as they can often be implemented as a simple rule systems in the client device; b) They tend to have lower computational cost than machine learning-based models; c) They are more straightforward to communicate to non-analytics decision makers and thus obtain organizational acceptance for them. However, heuristic-based rules eschew detailed predictions at the individual level. This often makes them more robust for predictions in vastly different environments, but leads to a loss of granular predictive ability in stable environments \cite{chintagunta2011-DCM,Goldstein2011-FAF}. Here we benchmark simple predictive heuristics against machine-learning models for rapid prediction of retaining players in a stable environment.

\section{Contribution}
Here the feasibility of predicting retention in F2P mobile games based on very short-term user behavior (i.e. as soon as possible after game download by the player) is evaluated. Retention prediction models are developed using a number of machine-learning models covering different windows of observation. These are benchmarked against a heuristic model developed using Decision Trees. Models are built based on a dataset of 130,000 players of the large mobile F2P game Jelly Splash. The dataset covers over 15 million sessions from the first 90 days of activity for a single cohort of users who installed the game over a seven day period. Accuracy varies across the observation window: Gameplay data covering a single session have minimal predictive power. Extending the observational period to the first day of gameplay slightly improves predictive accuracy, and finally using a one-week window substantially improves predictive ability of the models (accuracy 0.785-792). All three models exhibit similar accuracies across the feature windows suggesting that the advantage of modeling nonlinear relationships is limited. The accuracies of the models exceed those of a simple heuristic-based Decision Tree-model, but not substantially. This indicates that there is potential in using heuristic models for rapid, affordable and robust client-side predictions in F2P games. 

\section{Related work}
Due to space constraints the focus in this section will be on work directly related to the approaches used here: Churn models have been developed across a number of ICT sectors such as wireless communication, banking and insurance. In games, previous work on forecasting player behavior has focused on either Massively Multi-Player Online Games (MMOGs) or F2P mobile games. There have been very few cross-games studies, with exceptions including \cite{pittman2010-CVP}, who examined two MMOGs, and \cite{Sifa2014-TPP} who examined playtime patterns across more than 3000 titles. The methods that have been utilized vary from historical analysis, simple forecasting and multiple regression to machine learning techniques. The latter notably includes Decision Trees, Random Forest, Support Vector Machines, Neural Networks and Hidden Markov Models \cite{Sifa2015-PPD,Runge2014-CPH,Hadiji2014-PPC,thawonmas2011-ARO,yang2014-ipc,Xie2015-PPD}. In the latter context, previous work has mainly concentrated on churn prediction \cite{Runge2014-CPH,Hadiji2014-PPC} or predicting purchase decisions \cite{Sifa2015-PPD,Xie2015-PPD}. \cite{Hadiji2014-PPC} introduced different view-points to study the concept of churn and training classifiers to detect churn that is defined as a binary classification problem. The authors defined the concept of hard- and soft-churn, provide two different data generation methods to train any classification model and showed important factors for churning behavior in five different mobile free-to-play games. Similarly, \cite{Runge2014-CPH} predicted the departure of high value players in two casual social F2P games by comparing the performance of different classifiers and feature sets. Together with a supervised model for engagement modeling, \cite{Xie2015-PPD} concentrates on predicting first purchase in two social games using different classifiers. Finally, \cite{Sifa2015-PPD} particularly concentrate on predicting future purchase activities of players by formulating the process as a combination of a classification and a regression problem. The authors also emphasize the presence of rarity when analyzing premium players and provide a synthetic oversampling solution to predict rare purchase decisions. Across related work on F2P-based churn prediction, the importance of temporal features has been highlighted, e.g. features associated with the number of sessions per time period, the time between sessions, and average duration of sessions. Features related to specific game design were generally reported to be less important. 

Unlike previous work, the focus here is on the problem of rapid prediction of \emph{retaining} players by considering heuristic approaches owing to their ease of implementation and interpretation. Heuristics are strategies derived from experience with similar problems, using readily accessible information to control problem solving. They can be likened to rules of thumbs. They are often associated with the concept of satisfacing from economic decision-making \cite{Gigerenzer2009-HHW}. When finding an optimal solution is impossible or impractical, heuristic methods can be used for a satisfactory solution. They are used in a similar fashion in computer science, when the computational burden of complex methods is excessive. \cite{Goldstein2011-FAF} present a comprehensive review of their use in forecasting and prediction. \cite{Wubben-ICB}) empirically investigate their viability for use in database marketing. \cite{Artinger2015-HAD} detail their application in management more broadly. The work presented here can be viewed as a special case and extension of the previous authors' contributions.

\section{Definitions: retention and associated terms}
This paper operationalizes short-term retention prediction as a binary classification task: each player will be classified as either retained (1) or churned (0) by both our heuristic-based decision rules and comparison machine learning models. We define retention as having any game activity during the second week of game exposure. More specifically, a player will be labeled as retained if and only if he/she registers at least one game round in the period 7-14 days following installation. Examining the player’s second week of game exposure has several benefits: it facilitates the identification of engaged players while taking into account possible seasonal patterns in play (e.g. weekday vs. weekend); it minimizes confounding instances of disengaged players registering a single round long after they have stopped playing regularly; it enables training models and generating initial predictions shortly after launch, when the number of new players is highest and retention predictions most useful.

With respect to the single response defined above, we examine several different prediction periods and classification strategies. Each of our classifiers generate retention predictions using a player’s game activity from his/her installation date up until the end of one of three feature windows. A feature window is defined as an interval of time between the player's installation date and one of three cutoff points: 1) end of the player’s first session; 2) end of the player’s first day or: 3) end of the player’s first week. These feature windows represent periods of increasing game exposure and informational content. 

\begin{figure}[t!]
\centering
  \includegraphics[width=0.45\textwidth]{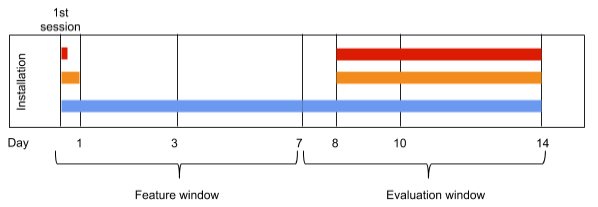}
  \caption{Feature- and evaluation windows used for model construction.}
\end{figure}

For each of these feature windows, three classification strategies are examined: i) Heuristic-based decision rules; 2) Several classifiers previously utilized for churn prediction; 3) An ensemble strategy combining the results of multiple classifiers. The goal here is to investigate the relationship between accuracy and actionability: observing more game activity yields more accurate predictions, but lowers the overall business value of these predictions as players who might have been incentivized to remain engaged will have already churned \cite{Runge2014-CPH,Sifa2015-PPD,Hadiji2014-PPC,rothenbuehler2015-HMM}. In addition, training traditional classifiers requires staff with specialized skills/knowledge, the transmission of user data to and from a central location, and an initial data collection period. In contrast, simple heuristic-based approaches can be deployed immediately after launch on the client devices themselves, and require little to no upkeep/monitoring after deployment. However, they are only useful if sufficiently accurate \cite{Wubben-ICB}. 

\section{Method and approach}

\subsection{Data and pre-processing}
Data for this analysis were provided by Wooga, are fully anonymized and notably contain installation, session- and rounds played data for a single cohort of users who installed the game over a seven day period in 2014. The data are from the game Jelly Splash on Apple's iOS platform. We observe all game sessions within the first year of exposure as well as all game rounds within the first 90 days for this single cohort of users. It is important to note that a session corresponds to a unique instance of a player opening the application on his or her device, while all actual game play occurs within rounds. It is possible for a player to record a session with no rounds, but all rounds must occur within sessions. In the dataset, 137,397 players installed the game, and 137,244 (or 99.9\%) of these players recorded a session (i.e. opened the game on their device) at some point. Of these players, only 94.5\% recorded at least one round (i.e. actually played the game). We restrict analysis to users registering a game session within the first seven days after installation and playing at least one game round during that first session. These sample restrictions preclude the confounding effects of individuals who install the game but never play, while also allowing for a common sample across our three feature windows. These restrictions reduce our sample size down to roughly 112,000 users. A small number of records with illogical timestamps and/or data values were further excluded prior to feature creation and analysis.

\begin{figure}[t!]
\centering
  \includegraphics[width=0.35\textwidth]{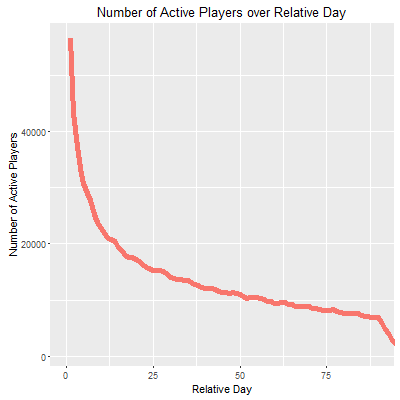}
  \caption{Number of active players on relative day since date of game installation).}
\end{figure}

\subsection{Feature definition and engineering}

The creation of features that adequately capture user characteristics and behavior is one of the most important aspects of any classification task. We did not have access to in-app purchase or player spending data, so our 18 created features primarily represent installation information and gameplay patterns. Many commonly used measures in the churn prediction literature are represented as well as several game-specific metrics relevant to our data set. Installation measures include the user’s device type (e.g. phone, tablet), geographic location, and whether or not he/she was referred from a marketing effort (acquired). Gameplay measures focus on play time (total days, total sessions, total rounds, average session duration, average round duration, total elapsed play time), intersession measures (current absence time relative to the end of the feature window, average time between sessions), social interaction (connected friends, player interaction), and round-specific statistics (average moves, average stars, maximum level). Installation-based measures are common across all three feature windows, whereas separate versions of each gameplay measure were created using only the sessions and rounds falling within each feature window. 

\begin{figure}[t!]
\centering
\includegraphics[width=0.35\textwidth]{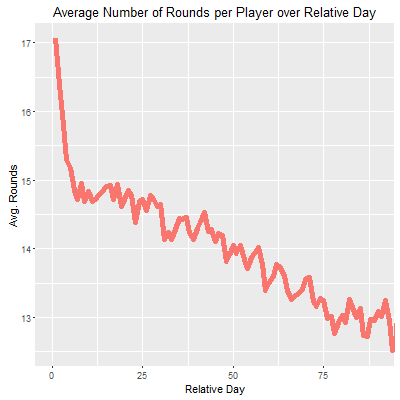}
 \caption{Average number of rounds played per player over relative day since date of game installation.}
\end{figure}

\subsection{Heuristic model development}


We explore short-term heuristics to rapidly predict whether a player will be retained days after game installation using simple decision trees. 10-fold cross validation was used to examine performance of heuristics based on gameplay data from the first session, day and week. The size of the tree was limited to keep the number of decision rules in each heuristic to 3 or 4. Results show that a day’s worth of player information can determine playing behavior a week or more into the future with decent accuracy. Multiple combinations of feature and evaluation windows were tested to investigate the trade-off between data collection times and heuristic performance.

The key variables used in the 1-day heuristic decisions are number of rounds, current absence time, and maximum level reached. The splits from the tree intuitively show that absence time of more than 20 hours after installation is a reliable determinant of player churn.

\begin{table*}[t!]
\centering
\label{}
\resizebox{0.60\textwidth}{!}{%
\begin{tabular}{llllll}
Feature Window & Evaluation Window & Accuracy & Precision & Recall & F1    \\
1 session      & 8 - 14 days        & 0.613            & 0.555     & 0.228  & 0.323 \\
1 day          & 8 - 14 days       & 0.686            & 0.639     & 0.509  & 0.567 \\
1 day          & 2 - 8 days        & 0.703            & 0.756     & 0.738  & 0.747 \\
1-3 days       & 4 - 10 days       & 0.747            & 0.787     & 0.681  & 0.73  \\
1-7 days       & 8 - 14 days       & 0.786            & 0.785     & 0.651  & 0.712
\end{tabular}%
}
\caption{Overview of feature- and evaluation windows and the prediction results for each.}
\end{table*}

To evaluate the robustness of the heuristics, we used an empirical approach to investigate the sensitivity of the heuristic's accuracy with respect to different training data. First, the entire data set is split into ten separate chunks (i.e. mutually exclusive random samples), with one reserved as a test set, and the remaining nine used for training samples. Decision trees are then trained separately on each of the nine chunks and tested on the hold-out sample.

We were also interested in whether our decision trees can correctly classify users with playing behaviors similar to those in the hold-out sample. To this end, we introduced "perturbations" to the hold-out sample by mapping each of its users to his/her nearest neighbor (in our feature space) outside of the hold-out sample with the same class label. (We chose to use this "perturbation" method in order not to make any assumptions about the smoothness of the underlying probability distributions of the classes, churned or retained.)

As an extension, we also investigated whether our decision trees can correctly classify users with playing behavior \textit{increasingly dissimilar} to those of players in the hold-out sample. To this end, we mapped the players to not just his/her nearest neighbor as before, but also to the player's \textit{i\textsuperscript{th}} (with 0 < \textit{i} < 10, referred to as the perturbation level below) nearest neighbor in the feature space. 

Comparing the results from each training chunk indicated that the performance of the decision trees is not sensitive to changes in the training data; our evidence from the second part of this empirical investigation also suggests that the performance in our decision trees is not particularly sensitive to perturbations in our hold-out sample. The range, mean, and standard deviation of misclassification rates at each perturbation level for a sample run is given in table 2 below.

\begin{table*}[t!]
\centering
\label{}
\resizebox{0.65\textwidth}{!}{
\begin{tabular}{lllllllllll}
& Holdout & 1-NN & 2-NN & 3-NN & 4-NN & 5-NN & 6-NN & 7-NN & 8-NN & 9-NN\\
Minimum & 0.317 & 0.31 & 0.308 & 0.312 & 0.31 & 0.308 & 0.307 & 0.309 & 0.31 & 0.308 \\
Maximum & 0.324 & 0.316 & 0.314 & 0.318 & 0.314 & 0.314 & 0.315 & 0.316 & 0.315 & 0.315\\
Mean & 0.32 & 0.313 & 0.312 & 0.314 & 0.312 & 0.311 & 0.31 & 0.312 & 0.312 & 0.312 \\
Standard Deviation & 0.002 & 0.002 & 0.002 & 0.002 & 0.001 & 0.002 & 0.002 & 0.003 & 0.002 & 0.002
\end{tabular}
}
\caption{Minimum, maximum, mean, and standard deviation of misclassification rates for 1-day heuristic (evaluation window 8-14 days) for each perturbation level.}
\end{table*}

\begin{figure}[t!]
\centering
\includegraphics[width=0.35\textwidth]{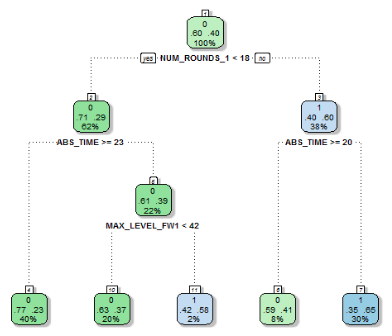}
 \caption{One day heuristic based on Decision Tree model.}
\end{figure} 

\subsection{Short-term prediction model development}

In this section we present an experimental evaluation of our churn prediction method using three popular machine learning classifiers for each feature window under study. The features relevant to each time period discussed above are used to train each classifier and predict whether or not a player will be retained in the second week after he/she installs the game. We compare the results of Logistic Regression (LR), Support Vector Machines (SVM), and Random Forest (RF) to assess the relative strengths and weaknesses of the different algorithms with respect to our specific data set, three feature windows, and prediction task. We report only key methodological steps and findings as an in-depth discussion of the classifiers themselves is beyond the scope of this work.

Included raw predictors, two-way interactions, and functional forms for all LR models were initially derived via an AIC-based stepwise search procedure and then fine-tuned by hand using 10-fold cross-validation error to compare candidate models. Hyperparameters for SVM (kernel, cost, gamma) and RF (variables per split, number of trees) models were tuned using a grid search method assessing candidate models with 10-fold cross-validation error. The data were randomly subsampled down to 10,000 observations for tuning to accommodate larger grid sizes and additional candidate comparisons under a reasonable amount of time and resources. We evaluated the relative and absolute performance of each classifier using 10-fold cross validation on the full data set. The same cross validation partitions were used for each of our three models to facilitate fair comparisons between the different classifiers. In addition, we also examined the performance of a simple majority-vote ensemble of the three models to assess the extent to which weaknesses in a single model could be overcome by strengths in the other two.

As the raw class distribution is 40.5\% retained, 59.5\% represents the naive baseline accuracy of class-weighted random predictions. With that baseline in mind, we see the models using only a single session of gameplay have little predictive power. Model accuracy improves slightly when using the first day of game activity, and substantially when taking into account the first week. It’s interesting to see that all three models exhibit similar overall accuracy with respect to each feature window, suggesting there may not be a large advantage to modeling nonlinear relationships in our data. However, important differences in the precision/recall trade-off exist across different model types, with the LR models typically exhibiting lower precision and higher recall than the SVM models. The majority-vote ensemble is the best overall performer, but adds little value over any one component model due to the similarity of all three. 

\section{Analysis and discussion}

\subsection{Model comparison}

While the accuracies of the three machine learning algorithms generally exceed those of the simple heuristic-based decision trees, the performance difference between the two approaches is not substantial. With respect to the single-session feature window the best machine learning algorithm outperformed the simple heuristic tree by only 1.2 percentage points of accuracy and had an F-1 score only 0.009 higher. For the single-day window the difference was even smaller: 0.3 percentage points of accuracy and an F-1 difference of 0.001. Lastly, using a full week of information the best machine learning algorithm improved accuracy over the heuristic by 0.6 percentage points and yielded an F-1 difference of 0.015. These results indicate that simpler decision rules implemented client-side can be possible for short-term retention prediction in mobile games.

The predictive power of our models falls generally within the range reported by the relevant literature. Looking at the results pertaining to the retention and feature window definitions most closely resembling those used in our experiments, \cite{Hadiji2014-PPC} arrive at retention F-1 scores ranging from 0.682 to 0.880 for five different F2P games. The authors use similar machine learning algorithms, but importantly have access to player purchase behavior to augment feature engineering. \cite{rothenbuehler2015-HMM} examine a 7-day moving average feature window with a similar retention definition and arrive at AUC values ranging from 79.1 to 79.6 for Neural Net and SVM models. These authors restrict features to generic session data (i.e. do not look at game-specific measures). Calculating the AUC of our 7-day feature window supervised learning ensemble model yields a value of 77.4, very close to the above-mentioned result. Some caution should be taken in comparing these results directly: each paper defines churn/retention uniquely, uses slightly different feature windows, and analyzes a different set of mobile games. 

\subsection{Feature importance}
Understanding the relationships between specific predictors and retention likelihoods helps inform intervention targeting. Towards this, we examine which player characteristics are most strongly related to retention overall and in each feature window. We evaluate these relationships using pairwise predictor-response correlations, logistic regression coefficient and standard error values, and random forest variable importance plots to assess the strength, size, and direction of each relationship.

Total Rounds and Total Playtime have the strongest overall effect on retention for the single-session feature window. Additionally, Average Stars surprisingly has a significant negative relationship with retention. We see a positive relationship for Average Duration and Average Moves, and retention rates also vary by Install Device Type: users installing the game on tablets generally exhibit lower retention relative to phone installations. Despite information from only the player’s first session not having much predictive power, the relationships above appear mostly intuitive: those who play longer immediately after installing the game are less likely to churn. 

Looking at the single-day and seven-day feature windows, overall playtime and consistent playtime are the main determinants of retention. Total Rounds, Total Sessions, and Average Duration are the strongest positive predictors, whereas Current Absence Time, Average Stars, and Average Time Between Sessions are the strongest negative factors. For the seven day feature window Current Absence Time becomes by far the strongest predictor, dominating regression models and random forest variable importance plots. These results seem to suggest a large number of players churn very soon after installing the game, whereas those who play for longer and over a more consistent basis in the first week are much more likely to be retained in the second. These findings are largely consistent with the wider literature. Another interesting finding is that measures related to skill (lower Average Moves, higher Average Stars) are actually inversely related to retention likelihood. This could represent certain players finding the game’s initial levels too easy and quickly losing interest. However, the fact that later levels are more difficult and require more moves on average may confound this relationship as players who immediately lose interest for any reason are unlikely to ever attempt these higher levels.

\begin{figure}[t!]
\centering
\includegraphics[width=0.40\textwidth]{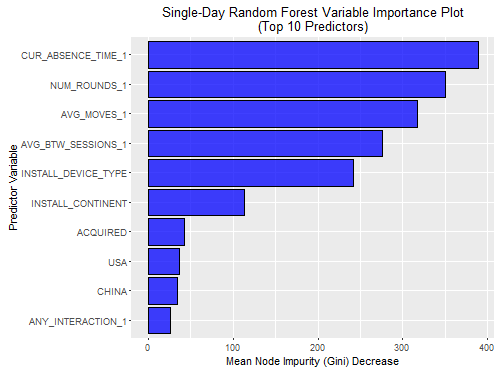}
 \caption{Example of feature importance results, here for the single-day random forest model. Note current absence time and number of rounds played as the most important features.}
\end{figure}

\subsection{Ability to identify long-term users}

In addition to identifying those users likely to churn rapidly after installing the game, these modeling techniques can also be used to identify long-term, potentially high-value customers. Identifying these customers and delivering targeted monetization strategies may be as or more important than knowing which users are likely to leave soon after installing the game, as an overwhelming proportion of F2P in-app purchases are generated by a very small proportion of players \cite{Sifa2015-PPD,Runge2014-CPH}. To approximately identify these long-term and potentially high-value players we look at 60-day retention, or whether the user registers a game round in the period 60-67 days after he/she installs the game. Although we cannot observe player spending directly with our available data, this long-term retention measure provides a simple definition of those players who are consistently engaged and likely to yield the highest ROI with respect to any targeted interventions.

In our analysis sample 15.2\% of players are categorized as long-term retained using the above definition. When looking at the results from our single-day models 27.1\% of those users predicted as short-term retained continue to play regularly past 60 days of game exposure. For the seven-day models 31.2\% those players predicted to be short-term retained meet the definition of long-term retention. While these percentages may seem low in an absolute sense, it is useful to benchmark them against the percentage of actual short-term retained players who continue on to be long-term retained. Of players categorized as short-term retained only 30.9\% are additionally categorized as long-term retained, implying the predictions from the short term models are actually slightly more accurate at identifying long-term players than the short-term class labels themselves. In essence, identifying long-term and potentially high-value players using only the first week of game exposure is a  difficult problem.

\begin{table}[t!]
\small
\centering
\label{}
\resizebox{0.47\textwidth}{!}{%
\begin{tabular}{cccccc}
Feature Window & Modeling Method & Accuracy & Precision & Recall & F1\\
Single Session & LR              & 0.623            & 0.58      & 0.21   & 0.308    \\
               & SVM             & 0.621            & 0.589     & 0.173  & 0.267    \\
               & RF              & 0.625            & 0.577     & 0.233  & 0.332    \\
               & ENSEMBLE        & 0.625            & 0.596     & 0.197  & 0.296    \\
First Day      & LR              & 0.684            & 0.641     & 0.505  & 0.565    \\
               & SVM             & 0.688            & 0.659     & 0.48   & 0.555    \\
               & RF              & 0.683            & 0.634     & 0.515  & 0.568    \\
               & ENSEMBLE        & 0.689            & 0.655     & 0.492  & 0.562    \\
First Week     & LR              & 0.785            & 0.741     & 0.713  & 0.727    \\
               & RF              & 0.789            & 0.776     & 0.666  & 0.717    \\
               & SVM             & 0.791            & 0.789     & 0.655  & 0.716    \\
               & ENSEMBLE        & 0.792            & 0.755     & 0.677  & 0.714   
\end{tabular}%
}
\caption{Results of evaluation of the relative and absolute performance of each classifer using 10-fold cross validation, across three models, as well as for the majority-vote ensemble of the three models.}
\end{table}

\section{Conclusion}
Previous work on churn prediction in games has generally focused on mid-length observation and prediction windows. e.g. 3-14 days of observation with prediction windows 7-14 days into the future \cite{Sifa2015-PPD,Runge2014-CPH,Hadiji2014-PPC,Xie2015-PPD}. However, in many F2P games there is substantial churn happening in the very beginning of the gameplay, meaning that the sooner prediction models can be build, the more designers (and educators) can proactively incentivize players to remain active. Prediction is equally interesting in a commercial context as well as from the perspective of human motivational- and attentional research. 

Here, the feasibility of rapid prediction of player retention in mobile F2P games is investigated, with multiple machine learning models applied across different windows of observation for comparison. The models exhibit similar accuracies within observation windows. This suggests that modeling of non-linear relationships only yields limited benefits. The accuracies of the models vary as a function of the observation window, increasing with length of the observation period. A further focus of the work presented here has been the introduction of the concept of heuristic models to prediction of player behavior. It can be concluded that the accuracies of the three advanced classifiers exceed those of a simple heuristic derived from a Decision Tree-model, but not substantially. This indicates that retaining players can be successfully determined with a short history of behavioral information and using heuristic prediction approaches. Finally, it suggests that a large part of the value of advanced analytics in games can potentially be accessed by relying on static heuristic models. They are beneficial in being robust, understandable and easy to deploy and scale.


\bibliographystyle{aaai}
\bibliography{bibliography}

\end{document}